\documentclass{article}

\usepackage[preprint,nonatbib]{neurips_2021}
\usepackage{multirow}

\usepackage[utf8]{inputenc} 
\usepackage[T1]{fontenc}    
\usepackage{hyperref}       
\usepackage{url}            
\usepackage{booktabs}       
\usepackage{amsfonts}       
\usepackage{nicefrac}       
\usepackage{microtype}      
\usepackage{xcolor}         
\usepackage{graphicx}
\title{Human Reaction Intensity Estimation with Ensemble of Multi-task Networks}

\author{%
  JiYeon Oh${^2}$, Daun Kim${^1}$, Jae-Yeop Jeong${^1}$, Yeong-Gi Hong${^1}$, and Jin-Woo Jeong${^1}$\\
  Department of Data Science${^1}$, Division of IISE${^2}$ \\
  Seoul National University of Science and Technology \\
  Seoul, Korea \\
  \texttt{\{dhwldus0906, daun, jaey.jeong, yghong, jinw.jeong\}@seoultech.ac.kr} \\
}

\begin{document}

\maketitle

\begin{abstract}
Facial expression in-the-wild is essential for various interactive computing domains. Especially, "Emotional Reaction Intensity" (ERI) is an important topic in the facial expression recognition task. In this paper, we propose a multi-emotional task learning-based approach and present preliminary results for the ERI challenge introduced in the 5th affective behavior analysis in-the-wild (ABAW) competition. Our method achieved the mean PCC score of 0.3254.
  
\end{abstract}

\section{Introduction}
Affective computing is a long-established area of interactive computing and one of the most active research areas in human computer interaction field, such as psychotherapy \cite{Khanna2022-rw}, game \cite{Setiono2021-hb}, and social robots \cite{Tian2022-uz}. In particular, accurately recognizing human response is crucial for HCI applications. Generally, the human response can be represented as emotion, which is explicit/implicit feedback on their internal states. Therefore, numerous studies have explored tracking emotion with various modalities, such as facial images \cite{Revina2018-lc}, speeches \cite{Khalil2019-bw}, and so on. Previous studies have shown that deep learning models perform well for relatively simple emotion recognition tasks, such as categorical emotion classification and valence/arousal regression \cite{Jeong2022-th, Kim2022-av, kollias2019face, kollias2021affect, kollias2021distribution}. However, these results do not necessarily imply that they reflect the overall state of human reactions. Therefore, more fine-grained tasks and solutions have been required to capture the wide range of internal human states and their corresponding responses. To address this challenge, more complex problems must be tackled, such as recognizing compound expressions \cite{Du2015-pm} and human reactions \cite{christ2022muse}. 
 
 As one of many attempts to address this issue, the 5th competition on Affective Behavior Analysis in-the-wild (ABAW)is held in conjunction with the Conference on Computer Vision and Pattern Recognition (CVPR) 2023 \cite{Kollias2023-ur}. 
 The ABAW competition \cite{kollias2020analysing, kollias2021analysing, kollias2022abawcvpr, kollias2022abaweccv} aims to ensure the entire feasibility of in-the-wild affective behavior analysis systems that can withstand video recording conditions, different situations, and display timing, regardless of human age, gender, race, and status. The 5th ABAW competition has the following tracks: 1) Valence-Arousal (VA)  2) Expression recognition (EXPR) 3) Action Unit detection (AU) 4) Emotional Reaction Intensity (ERI). The first three tracks are based on the Aff-Wild2 database \cite{kollias2019expression}, which is an extension of the Aff-wild database \cite{zafeiriou2017aff, kollias2019deep}. The last track is based on the Hume-Reaction dataset.
  
In this work, we report our methods for the ERI Estimation challenge and early results. For this challenge, the Hume-Reaction dataset \cite{christ2022muse} is employed for model train and validation. It consists of subjects responding to a more fine-grained range of various emotional video-based stimuli. It is multi-modal (i.e., facial images and audio) and consists of approximately 75 hours of video recordings recorded via a webcam inside subjects’ homes. A total of 2,222 subjects were recorded across two cultures, South Africa and the United States. Each sample within the dataset has been self-annotated by the subjects themselves for the intensity of 7 emotional experiences in a range from 1-100: Adoration, Amusement, Anxiety, Disgust, Empathic Pain, Fear, and Surprise.

\section{Method}
\subsection{Overview}
The Hume-Reaction dataset contains approximately 75 hours of video recordings. In other words, leveraging the time-series image frames and audio characteristics of these videos is the key breakthrough for ERI. 
In this work, we use only a set of facial images extracted from videos in the Hume-Reaction dataset. To derive high-level performance, the following points were carefully taken into account in particular: better feature representation and handling time series context. Hence, we utilize a deep-learning model called MTL-DAN (refer to \cite{wen2021distract} for original DAN), which is a modified architecture to embrace robust feature representation in a multi-emotional task learning manner. In addition, to exploit the context of time-series data, we feed the final feature representation of MTL-DAN to a recurrent neural network, LSTM, \cite{Kratzert2021-ui} for the final estimation of emotional reaction intensity. For the ERI challenge, the mean Pearson Correlation Coefficient(PCC) across all 7 reaction categories was used as a metric. More details about our framework can be found in Section \ref{sec:emotion Feature Extractor} and \ref{sec:lstm}.

\begin{figure}[t]
    \centering
    \includegraphics[width=\columnwidth]{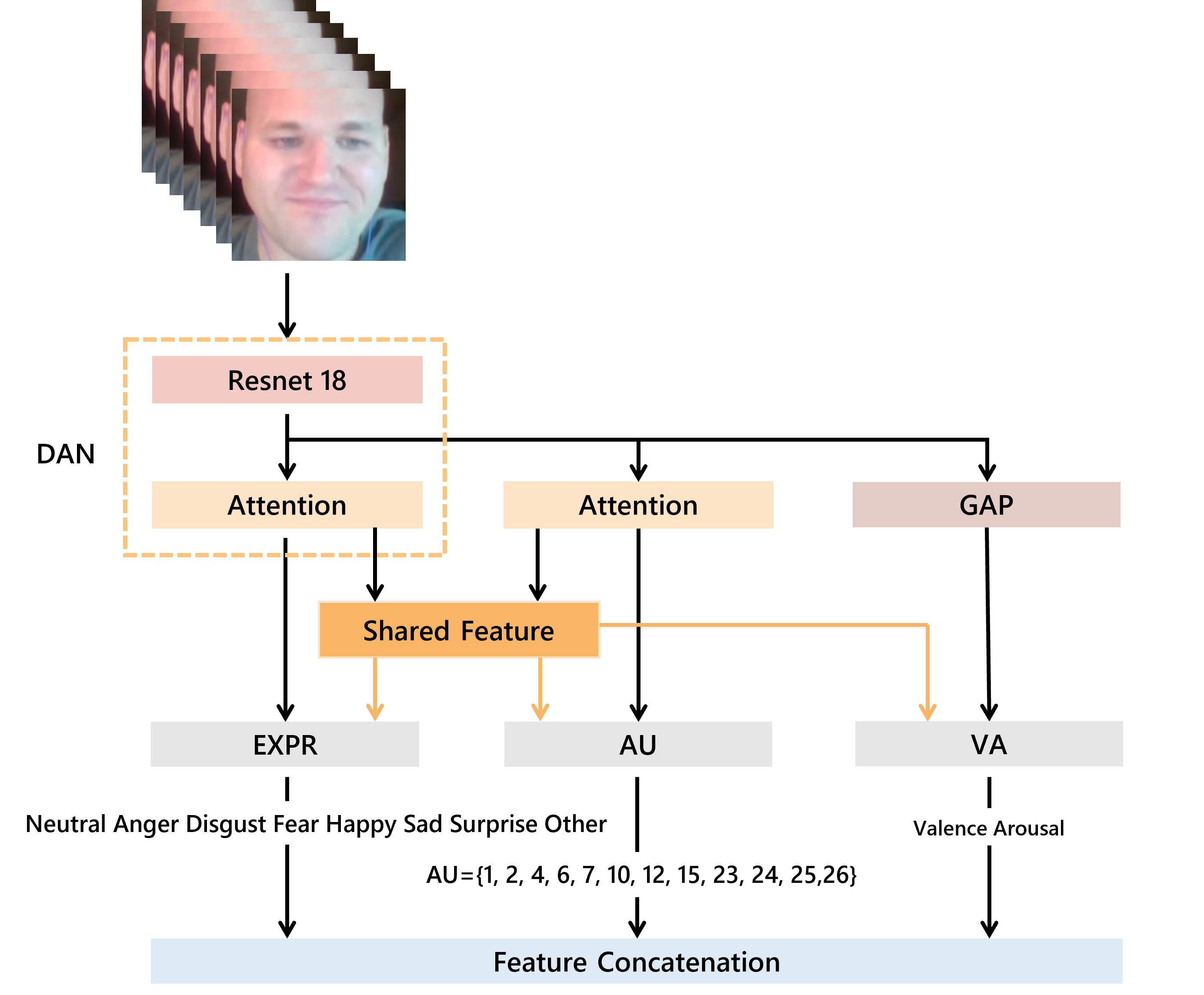}
    \caption{Architecture of MTL-DAN}
    \label{fig:MTL}
\end{figure}

\subsection{Training data}
\begin{figure}[t]
    \centering
    \includegraphics[width=\columnwidth]{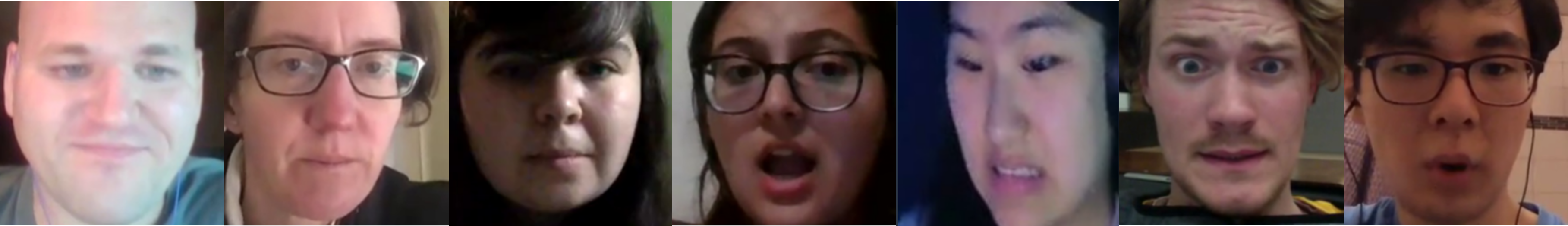}
    \caption{Training data used in our study}
    \label{fig:dataset}
\end{figure}

Figure \ref{fig:dataset} represents facial images extracted from videos in the Hume-Reaction dataset. We segment each video by extracting the first frame from every 30 frames. On average, each video is represented using 11.62 frames. 

\begin{figure}[t]
    \centering
    \includegraphics[width=\columnwidth]{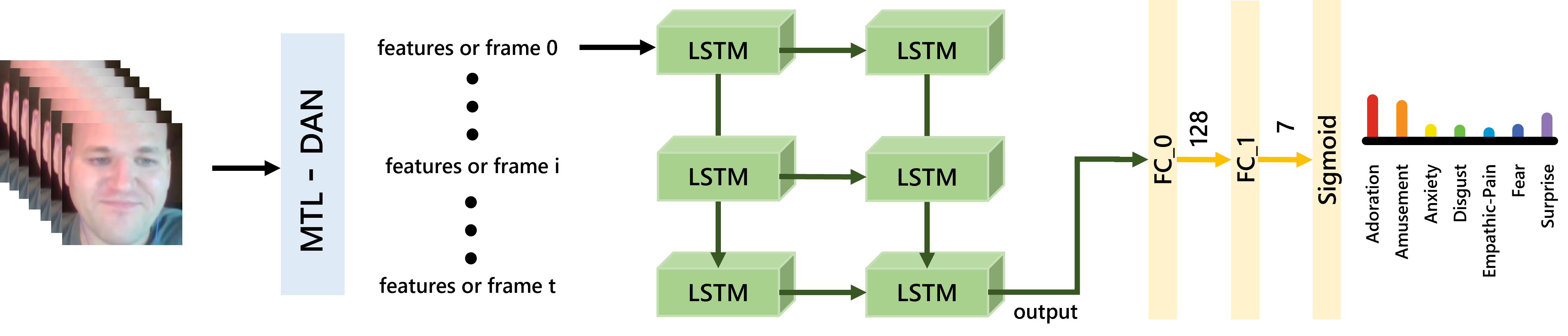}
    \caption{Overview of the architecture used in this study}
    \label{fig:architecture}
\end{figure}

\begin{figure}[t]
    \centering
    \includegraphics[scale=0.5]{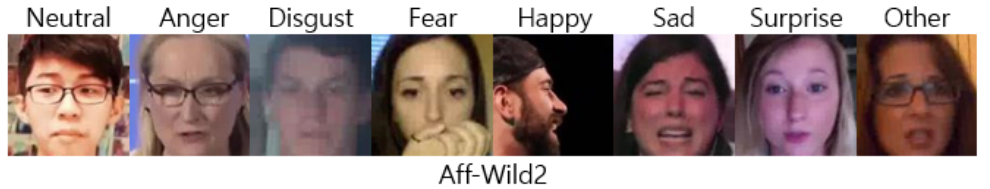}
    \caption{Example of training images for MTL-DAN}
    \label{fig:MTLDAN}
\end{figure}


\subsubsection{DAN Architecture for Multi-emotional task Learning}
\label{sec:emotion Feature Extractor}
As depicted in Figure \ref{fig:architecture}, we adopt a deep learning-based facial expression recognition approach called "DAN" \cite{wen2021distract} which is a high-performance method for the AffectNet database \cite{mollahosseini2017affectnet}. The DAN architecture has two phases: feature extractor and attention parts. In the attention phase, there are multi-head cross-attention units which consist of a combination of spatial and channel attention units. The DAN architecture used in our framework is a modified version, called MTL-DAN. We extend the original DAN architecture to MTL-DAN that can jointly optimize categorical expression recognition (EXPR), action unit detection(AU), and valence/arousal regression(VA) in a multi-task learning manner. More details of MTL-DAN are as follows. As shown in Fig. \ref{fig:MTL}, Resnet18 \cite{he2016deep} is a feature extractor in the original DAN that creates an image feature containing emotional attributes. 
In DAN-MTL, the features from ResNet18 is utilized differently to handle different emotional recognition tasks. First, feature representation from ResNet18 is not fed into an attention block in the case of VA task, while EXPR and AU go through an independent attention block. After that, we concatenate feature representation from the attention blocks of EXPR and AU, which is represented as a shared feature in Fig.\ref{fig:MTL} Finally, we feed together the shared feature and each independent feature into each classification/detection/regression(VA) head for robust feature representation. As a result, MTL-DAN generates the outputs of three types: $V_{EXPR} \in  R^{8}$, $V_{AU} \in  R^{12}$, and $V_{VA}\in  R^{2}$. 
In our experiments, we initialize the parameters of MTL-DAN with pretrained MTL-DAN using Aff-wild2 multi-task learning challenge dataset \cite{kollias2022abawcvpr}. Also, we do not update the parameters of MTL-DAN to prevent overfitting to emotional reaction labels.

\subsubsection{Regression Head}
\label{sec:lstm}
In this section, we describe how the estimation of emotional reaction intensity using a regression head (RH), which consists of a series of LSTM and fully connected layers, is made. Prior to feeding the outputs of MTL-DAN into LSTM, we concatenate three outputs from EXPR, AU, and VA heads.
Finally, our framework produces the final prediction of ERI estimation with the sigmoid function.

\section{Results}
All the experiments were conducted using a GPU server with six NVIDIA RTX 3090 GPUs, 128 GB RAM, Intel i9-10940X CPU, and Pytorch framework. 

The goal of the experiments is to measure the performance of the model in estimating the emotional reaction intensity. To improve the performance, we experimented with two loss functions: the Concordance Correlation Coefficient (CCC) loss and the Pearson Correlation Coefficient (PCC) loss. We selected the loss function that resulted in the best performance.
Our preliminary results on the official validation set for the ERI Estimation challenge was 0.3254 in terms of the mean PCC score, which outperforms baseline methods, such as ResNet50-FAU (0.2840) and ResNet50-VGGFace2 (0.2488) \cite{Kollias2023-ur}.

%


\section{Conclusion}

In this paper, we proposed a multi-emotional task learning-based architecture, MTL-DAN, and presented the preliminary results for the ERI estimation challenge in the 5th ABAW competition. Our method produced a mean PCC score of 0.3254 on the validation set for the ERI estimation challenge. The implementation details and validation results may be updated after the submission of this paper to arxiv.

\nocite{*}
{\small
\bibliographystyle{ieee_fullname}
\bibliography{egbib}
}
\end{document}